# scientific reports

# Simultaneous corn and soybean yield prediction from remote sensing data using deep transfer learning

Saeed Khaki[1][✉], Hieu Pham[2] & Lizhi Wang[1]

Large-scale crop yield estimation is, in part, made possible due to the availability of remote sensing data allowing for the continuous monitoring of crops throughout their growth cycle. Having this information allows stakeholders the ability to make real-time decisions to maximize yield potential. Although various models exist that predict yield from remote sensing data, there currently does not exist an approach that can estimate yield for multiple crops simultaneously, and thus leads to more accurate predictions. A model that predicts the yield of multiple crops and concurrently considers the interaction between multiple crop yields. We propose a new convolutional neural network model called YieldNet which utilizes a novel deep learning framework that uses transfer learning between corn and soybean yield predictions by sharing the weights of the backbone feature extractor. Additionally, to consider the multi-target response variable, we propose a new loss function. We conduct our experiment using data from 1132 counties for corn and 1076 counties for soybean across the United States. Numerical results demonstrate that our proposed method accurately predicts corn and soybean yield from one to four months before the harvest with an MAE being 8.74% and 8.70% of the average yield, respectively, and is competitive to other state-of-the-art approaches.

The use of satellites and other remote sensing mechanisms has proven to be vital in monitoring the growth of various crops both spatially and temporally[1]. Common uses of remote sensing are to extract various descriptive indices such as normalized difference vegetation index (NDVI), temperature condition index, enhanced vegetation index, and leaf area index[2]. This information can be used for drought monitoring, detecting excessive soil wetness, quantifying weather impacts on vegetation, the evaluation of vegetation health and productivity, and crop yield forecasting[3–5]. Moreover, a large-scale collection of environmental descriptors such as surface temperature and precipitation can also be identified from satellite images[6,7].

Accurate and non-destructive identification of crop yield throughout a crop's growth stage enables farmers, commercial breeding organizations, and government agencies to make decisions to maximize yield output, ultimately for a nation's economic benefit. Early estimations of yield at the field/farm level play a vital role in crop management in terms of site-specific decisions to maximize a crop's potential. Numerous approaches have been applied for crop yield prediction, such as manual surveys, agro-meteorological models, and remote sensing based methods[8]. However, given the size of farming operations, manual field surveys are neither efficient nor practical. Moreover, modeling approaches have limitations due to the difficulty in determining model parameters, the scalability to large areas, and the computational resources necessary[9].

Recently, satellite data has become widely available in various spatial, temporal, and spectral resolutions and can be used to predict crop yield over different geographical locations and scales[10]. The availability of Earth observation (EO) data has created new ways for efficient, large-scale agricultural mapping[11]. EO data enables a unique mechanism to capture crop information over large areas with regular updates to create maps of crop production and yield. Nevertheless, due to the high spatial resolution needed for accurate yield predictions, Unmanned Aerial Vehicles (UAV) have been promoted for data acquisition[12]. Although UAV platforms have demonstrated superior image capturing abilities, without the assistance of a large workforce, accurately measuring the yield for large regions is not feasible.

Indeed, advances in technology have made yield prediction more accessible and accurate both through machine learning and statistical approaches. Historically, common methods to predict crop yield include random forests, linear regression, and ensemble approaches[13–15]. However, recently crop yield predictions have been dominated by deep learning approaches. Recent works have applied multi-layer perceptrons to predict yield in wheat, corn, and strawberry yield by combining observed phenotypic data and environmental data[16–19]. Moreover, there is an increasing amount of literature combining convolutional neural networks and yield prediction from UAV imagery[20–22]. It is evident from recent works that no matter the data acquisition mechanism, there is an apparent shift in utilizing deep neural networks for crop yield predictions.

[1]Department of Industrial and Manufacturing Systems Engineering, Iowa State University, Ames, IA 50011, USA. [2]Syngenta Seeds, Slater, IA 50244, USA. ✉email: skhaki@iastate.edu

In regards to the scope of our paper, remote sensing capabilities have been extensively used for estimating crop yield around the world of various scales (field, county, state). Various studies have been performed using different vegetation indices to estimate yield in maize, wheat, grapes, rice, corn, and soybeans using random forests, neural networks, multiple linear regression, partial least squares regression, and crop models[23–31]. Similar to tabular yield data and UAV imagery, remote sensing yield predictions were historically dominated by traditionally machine learning and statistical approaches. However, there is a recent trend towards combining convolutional neural networks with satellite imagery for crop yield predictions[32–34]. These research papers showcase the potential, power, and accuracy deep learning and remote sensing have on estimating yield at a large scale.

For this paper, we consider two main crops in the Midwest United States (US), corn (*Zea mays* L.) and soybeans (*Glycine max*). According to the United States Department of Agriculture, in 2019, 89.7 million and 76.1 million acres of corn and soybean were planted, respectively. The combination of these two crops makes up approximately 20% of active US farmland[35]. Given the sheer size of these farming operations combined with the growing population, actions must be taken to maximize yield. With the help of remote sensing, government officials, as well as farmers, can monitor crop growth (at various scales) to ensure proper crop management decisions are made to maximize the yield potential.

Although approaches exist to estimate yield with various means, current models are limited in that they only estimate yield for a single crop. That is, there does not exist any model making use of remote sensing data to predict the yield of multiple crops simultaneously. More specifically, the scientific question we answer in this paper is determining how to extract transferable information about the yield prediction of one crop to be used for the yield prediction of another crop. Moreover, because individual models need to be constructed for each crop, long computational times prevent machine learning methods from being adopted over large areas[36,37]. One way to alleviate these issues is by considering simultaneous (multi-target) regression of yield. By predicting multiple response variables simultaneously, interaction effects are considered between response variables leading to improved accuracy, and computational performance is reduced due to having a single model predict multiple outputs as opposed to multiple models predicting multiple outputs.

Multi-target regression using support vector regression has been successful in estimating different biophysical parameters from remote sensing images outperforming single-target regression techniques both in terms of computational performance and accuracy[38]. Additionally, a multi-target Gaussian regressor created to identify canopy biomass in rice paddies was shown to obtain superior accuracy when compared to a single-target Gaussian regression[39]. Indeed, across a variety of domains and applications, simultaneous regression of multiple responses has shown promise with improved accuracy and computational performance[40–42].

Given the promising attributes of a multi-target regression model, we apply this approach to our remote sensing use case. Having such a model would, hopefully, result in more accurate predictions by considering the interactions between crops and estimating yields accordingly. Additionally, because predictions are simultaneous, the results are less computationally intensive than computing a model for each crop. Therefore, we propose a new deep learning framework called YieldNet utilizing a novel deep neural network architecture. This model makes use of a transfer learning methodology that acts as a bridge to share model weights for the backbone feature extractor. Moreover, because of the uniqueness associated with simultaneous yield prediction for two crops, we propose a new loss function that can handle multiple response variables. Specifically, the novelties of our approach include:

1. A new model that extracts transferable information from both corn and soybean to improve the yield prediction of both crops simultaneously.
2. The proposed method utilizes transfer learning through sharing weights of the backbone feature extractor among multiple crops, therefore, improving computational efficiency. To the best of our knowledge, this is the first deep learning approach combining multi-target regression, convolutional neural networks, and remote sensing data for crop yield prediction.
3. A new loss function is proposed to consider the multi-target response variable.
4. The weight sharing property of the proposed method substantially decreases the number of model parameters and subsequently helps the training process despite having limited labeled data.
5. The effectiveness of our proposed method is demonstrated on large-scale geospatial data from 1132 counties for corn and 1076 counties for soybean covering 13 states across the United States.

## Methodology

The goal of this paper is to simultaneously predict the average yield per unit area of two different crops (corn and soybean) both grown in regions of interest (US counties) based on a sequence of images taken by satellite before harvest. Let $I_{l,k}^t \in \mathbb{R}^{H \times W \times d}$ and $Y_{l,k}^c \in \mathbb{R}^+$ denote, respectively, the remotely sensed multi-spectral image taken at time $t \in \{1, \ldots, T\}$ during the growing season and the ground truth average yield of crop type $c \in \{1, 2\}$ planted in location $l$ at year $k$, where $H$ and $W$ are the image's height and width and $d$ is the number of bands (channels). Thus, the dataset can be represented as

$$D = \left\{ \left( \left( I_{l,k}^1, I_{l,k}^2, \ldots, I_{l,k}^T \right), \left( Y_{l,k}^1, Y_{l,k}^2 \right) \right) \Big| l \in \{1, \ldots, L\}, k \in \{1, \ldots, K\} \right\},$$

where $L$ and $K$ are the numbers of locations and years, respectively.

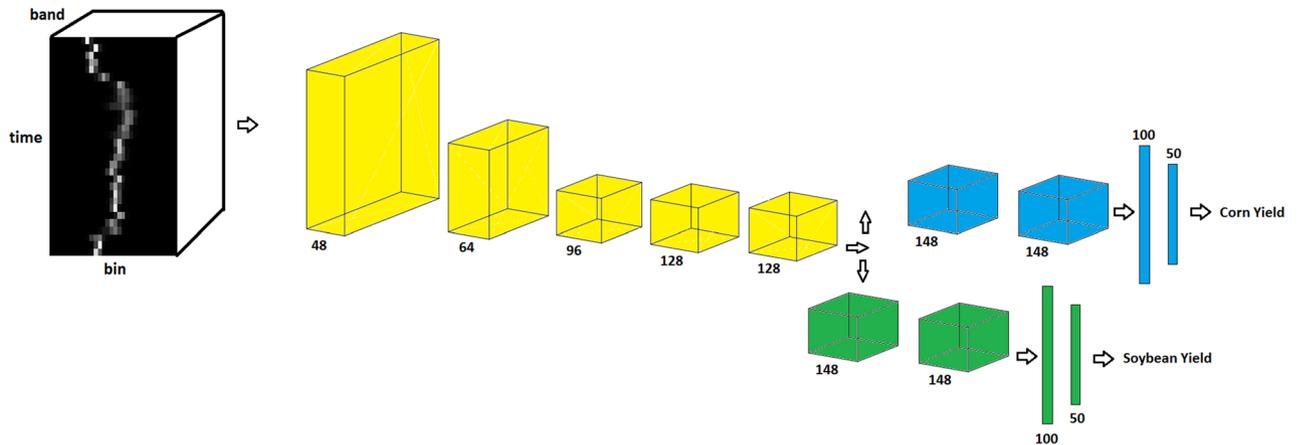

**Figure 1.** Outline of the YieldNet architecture. The input of the YieldNet is 3-D histograms $H \in \mathbb{R}^{T \times b \times d}$ and the 2-D convolution operation is performed over the 'time' and 'bin' dimensions while considering bands as channels. The number under feature maps indicates the number of channels. Convolutional layers denoted with yellow color work as a backbone feature extractor and share weights for both corn and soybean yield predictions.

For our approach, we do not use end-to-end training due to the main following reasons: (1) the number of labeled training data is limited, and (2) inability to use transfer learning from popular benchmark datasets such as Imagenet[43] due to domain difference and multi-spectral nature of satellite images. Therefore, we reduce the dimensionality of the raw remote sensing images under the permutation invariance assumption which states the average yield mostly depends on the number of different pixel types rather than the position of the pixels in images due to the infeasibility of end-to-end training. Similar approaches have been used in other studies[34,44]. As a result, we separately discretize the pixel values for each band of a multi-spectral image $I_{l,k}^t$ into $b$ bins and obtain a histogram representation $h_{l,k}^t \in \mathbb{R}^{b \times d}$. If we obtain histogram representations of the sequence of multi-spectral images denoted as $(h_{l,k}^1, h_{l,k}^2, \ldots, h_{l,k}^T)$ and concatenate them through time dimension, we can produce a compact histogram representations $H_{l,k} \in \mathbb{R}^{T \times b \times d}$ of the sequence of multi-spectral images taken during growing season. As such, the dataset D can be re-written as the following and we will use this notation throughout the rest of the paper:

$$D = \left\{ \left( H_{l,k}, (Y_{l,k}^1, Y_{l,k}^2) \right) \Big| l \in \{1, \ldots, L\}, k \in \{1, \ldots, K\} \right\}.$$

Given the above dataset, this paper proposes a deep learning based method, named YieldNet, that learns the desired mapping $H_{l,k} \in \mathbb{R}^{T \times b \times d} \mapsto (Y_{l,k}^1, Y_{l,k}^2)$ to predict the yields of two different crops simultaneously.

**Network architecture.** Crop yield is a highly complex trait that is affected by many factors such as environmental conditions and crop genotype which requires a complex model to reveal the functional relationship between these interactive factors and crop yield.

We propose a novel convolutional neural network architecture which is a highly non-linear and complex model. Convolutional neural networks belong to the class of representation learning methods which automatically extract necessary features from raw data without the need for any handcrafted features[45,46]. Figure 1 outlines the architecture of the proposed method. Table 1 shows the detailed architecture of the proposed model. We use a 2-D convolution operation which is performed over the 'time' and 'bin' dimensions while considering bands as channels. As such, convolution operation over the time dimension can help capture the temporal effect of satellite images collected over time intervals.

In our proposed network, the first five convolutional layers share weights for corn and soybean yield predictions which are denoted with yellow color in Fig. 1. These layers extract relevant features from input data for both corn and soybean yield predictions. The intuition behind using one common feature extractor is that it significantly decreases the number of parameters of the network, which helps training the model more efficiently given the scarcity of the labeled data. In addition, many low-level features captured by the comment feature extractor reflect general environmental conditions that are transferable between corn and soybean yields. All convolutional layers are followed by batch normalization[47] and ReLU nonlinearities in our proposed network. Batch normalization is used to accelerate the training process by reducing internal covariate shifts and regularizing the network. We use two convolutional layers in both corn and soybean heads which are followed by two fully connected layers.

| Type/stride | Padding | Filter size | Number of Filters |
|---|---|---|---|
| Conv/s2 | Valid | 7 × 7 | 48 |
| Conv/s2 | Valid | 5 × 5 | 64 |
| Conv/s2 | Same | 5 × 5 | 96 |
| Conv/s1 | Same | 3 × 3 | 128 |
| Conv/s1 | Same | 3 × 3 | 128 |
| **Corn head** | | | |
| Conv/s1 | Same | 3 × 3 | 148 |
| Conv/s1 | Same | 3 × 3 | 148 |
| FC-100 | | | |
| FC-50 | | | |
| **Soybean head** | | | |
| Conv/s1 | Same | 3 × 3 | 148 |
| Conv/s1 | Same | 3 × 3 | 148 |
| FC-100 | | | |
| FC-50 | | | |

**Table 1.** YieldNet architecture. The first five convolutional layers work as a backbone feature extractor and share weights for both corn and soybean yield predictions.

**Network loss.** To jointly minimize the prediction errors of corn and soybean yield forecasting, we propose the following loss function:

$$L = \max \left( \frac{1}{N_c} \sum_{i=1}^{N_c} \left( \frac{Y_i^c - \hat{Y}_i^c}{\bar{Y}^c} \right)^2, \frac{1}{N_s} \sum_{i=1}^{N_s} \left( \frac{Y_i^s - \hat{Y}_i^s}{\bar{Y}^s} \right)^2 \right) \quad (1)$$

where $Y_i^c, \hat{Y}_i^c, Y_i^s, \hat{Y}_i^s, N_c, N_s, \bar{Y}^c$, and $\bar{Y}^s$ denote $i$th average ground truth corn yield, $i$th predicted corn yield, $i$th average ground truth soybean yield, $i$th predicted soybean yield, number of corn samples, number of soybean samples, average corn yield, and average soybean yield, respectively. Our proposed loss function is a normalized Euclidean loss which makes the corn and soybean losses to have the same scale. We use the maximum function in our proposed loss function to make the training process more stable and ensure that both corn and soybean losses are optimized.

## Experiments and results

In this section, we present the dataset used in our study and then report the results of our proposed method along with other competing methods in corn and soybean yield predictions. We conducted all experiments in Tensorflow[48] on an NVIDIA Tesla V100 GPU.

**Data.** The data analyzed in this study included three sets: yield performance, satellite images, and cropland data layers.

- The yield performance dataset includes the observed county-level average yield for corn and soybean between 2004 and 2018 across 1132 counties for corn and 1076 for soybean within 13 states of the US Corn Belt: Indiana, Illinois, Iowa, Minnesota, Missouri, Nebraska, Kansas, North Dakota, South Dakota, Ohio, Kentucky, Michigan, and Wisconsin, where corn and soybean are considered the dominant crops[49]. Figures 2 and 4 depict the US Corn Belt and histogram of yields, respectively. The summary statistics of the corn and soybean yields are shown in Table 2.
- Satellite data contains MODIS products including MOD09A1 and MYD11A2. MOD09A1 product provides an estimate of the surface spectral reflectance of Terra MODIS bands 1-7 at 500m resolution and corrected for atmospheric conditions[50]. The MYD11A2 product provides an average land surface temperature which has the day and night-time surface temperature bands at 1km resolution[51]. It is worth noting that MOD09Q1 data exists at 250m resolution, however, it encompasses only 2 reflectance bands. Thus, although higher resolution, the data contains less information[52]. Figure 3 depicts an example multispectral image of land surface temperature and surface spectral reflectance for Adams county in Illinois. These satellite images were captured at 8-days intervals and we only use satellite images captured during growing seasons (March–October). As such, satellite images are collected 30 times a year in our study. We discretize all multispectral images using 32 bins to generate the 3-D histograms $H \in \mathbb{R}^{T \times b \times d}$, where $T = 30$, $b = 32$, and $d = 9$.
- USDA-NASS cropland data layers (CDL) is crop-specific land cover data that are produced annually for different crops based on moderate resolution satellite imagery and extensive agricultural ground truth[53]. In this paper, cropland data layers are used for both corn and soybean to focus on only croplands within each county and exclude non-croplands such as buildings and streets from satellite images.

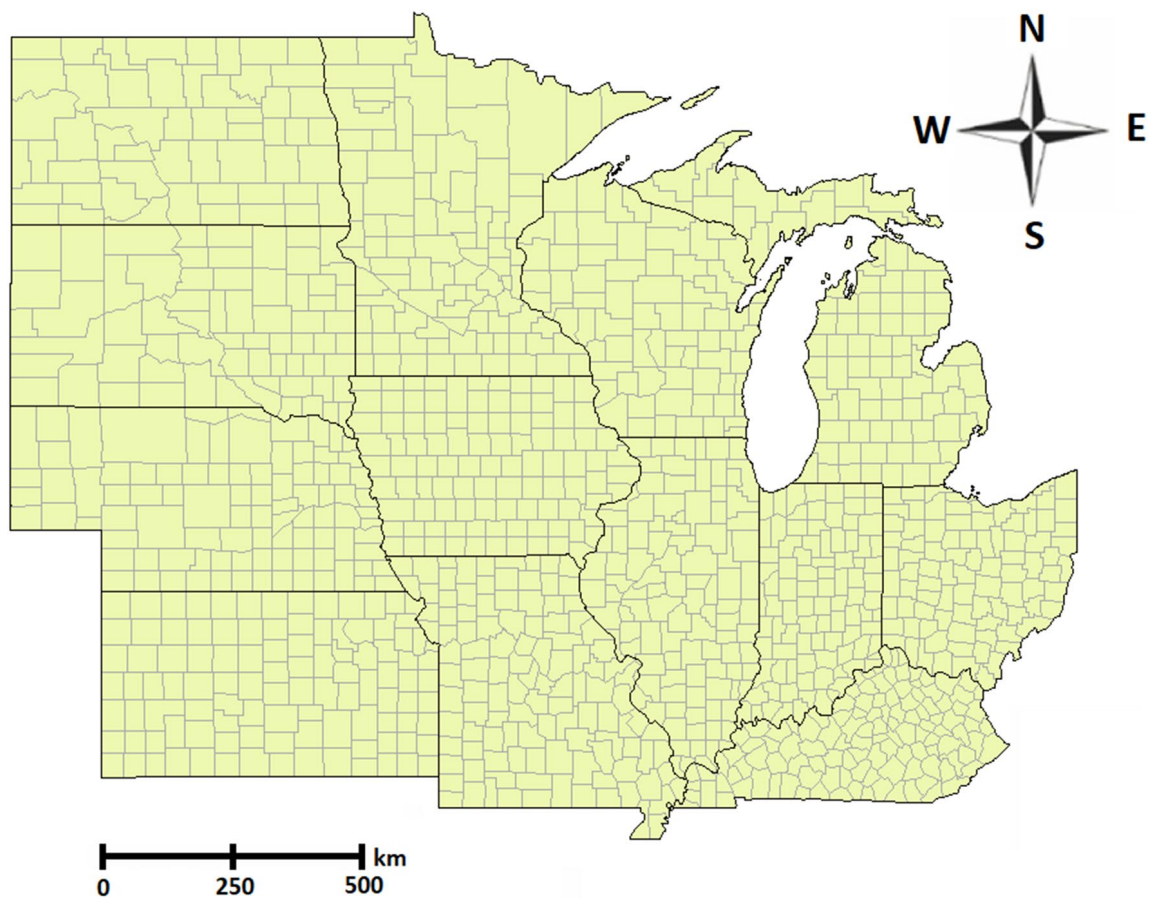

**Figure 2.** The map of US Corn Belt.

| Summary statistics | Corn | Soybean |
|---|---|---|
| Year range | 2004–2018 | 2004–2018 |
| Average yield | 146.68 | 45.02 |
| Standard deviation of yield | 36.03 | 10.08 |
| Minimum yield | 18.3 | 9.3 |
| Maximum yield | 246.7 | 82.3 |
| Number of locations | 1132 | 1076 |
| Number of observations | 13,992 | 12,502 |

**Table 2.** The summary statistics of yield data across all years. The unit of yield is bushels per acre.

**Design of experiments.** We compare our proposed method with the following models to evaluate the efficiency of our method.

*Random forest (RF)*[54]. RF is a non-parametric ensemble learning method that is robust against overfitting. We set the number and the maximum depth of trees in RF to be 150 and 20, respectively. We tried different numbers and the maximum depth of trees and found that these hyperparameters resulted in the most accurate predictions.

*Deep feed forward neural network (DFNN).* DFNN is a highly nonlinear model which stacks multiple fully connected layers to learn the underlying functional relationship between inputs and outputs. The DFNN model has nine hidden layers, each having 50 neurons. We used batch normalization for all layers. The ReLU activation function was used in all layers except the output layer. The model was trained for 120,000 iterations. Adam optimizer was used to minimize the Euclidean loss function.

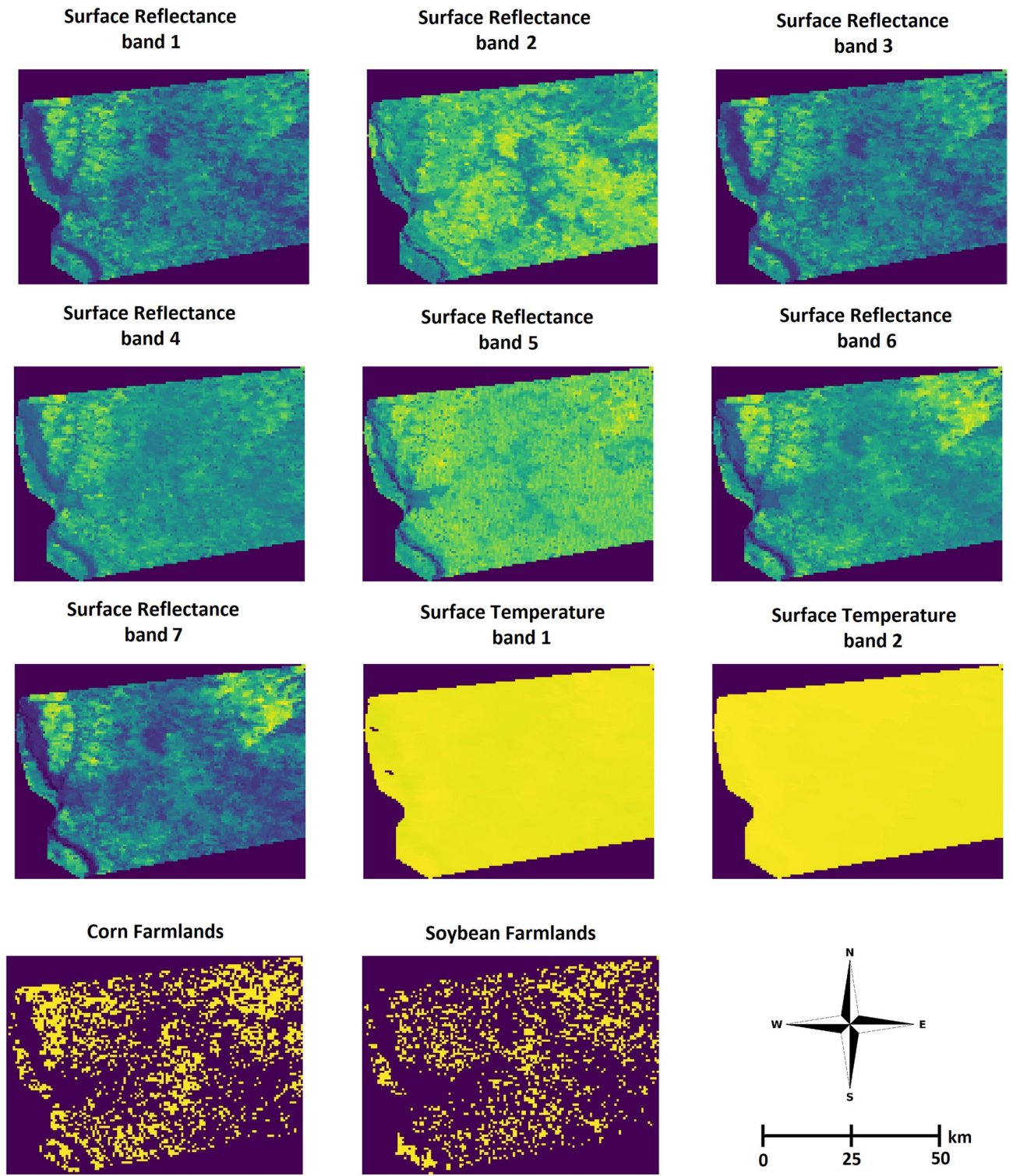

**Figure 3.** The multispectral images of surface spectral reflectance and land surface temperature for Adams county in Illinois captured on May 8th, 2011. Corn and soybean farmland images indicate cropland areas within the Adams county in Illinois in year 2011.

*3-dimensional convolutional neural network (3D-CNN).*    3D-CNN is a highly non-linear model which employs 3-dimensional kernels in its convolution operations to capture spatio-temporal features in the data[55]. The 3D-CNN captures the temporal effects of remote sensing data as well as spatial and intra-band features extracted from the individual images[56]. In this paper, we used a homogeneous network architecture for the 3D-CNN which is found effective in other studies[55,57]. Table 3 shows the network architecture of the 3D-CNN.

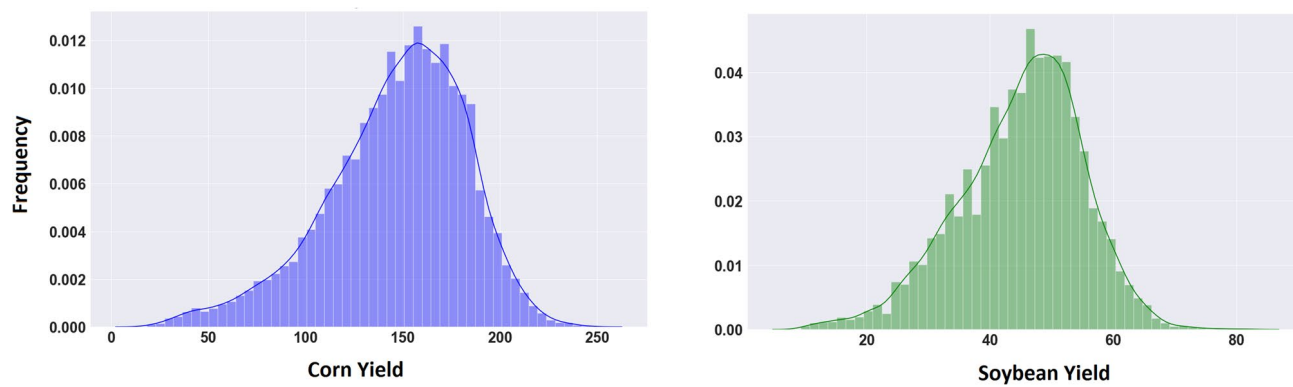

**Figure 4.** The histograms of the corn and soybean yields. The unit of yield is bushels per acre.

| Type/stride | Padding | Filter size | Number of filters |
|---|---|---|---|
| Conv/s1 | Same | 3 × 3 × 1 | 64 |
| Max pool | Same | 4 × 4 × 2 | – |
| Conv/s1 | Same | 3 × 3 × 1 | 64 |
| Conv/s1 | Same | 3 × 3 × 1 | 64 |
| Conv/s1 | Same | 3 × 3 × 1 | 64 |
| Conv/s1 | Same | 3 × 3 × 1 | 64 |
| Conv/s1 | Same | 3 × 3 × 1 | 128 |
| Max pool | Same | 2 × 2 × 1 | – |
| FC-256 | | | |
| FC-128 | | | |

**Table 3.** 3D-CNN network architecture. The input of the 3D-CNN is 3-D histograms $H \in \mathbb{R}^{T \times b \times d}$ and the 3-D convolution operation is performed over the 'time', 'bin', and 'band' dimensions.

*Regression tree (RT)*[58]. RT is a nonlinear model which does not make any assumption regarding the mapping function. We set the maximum depth of the regression tree to be 12 to decrease the overfitting.

*Lasso*[59]. Lasso adopts $L_1$ norm to reduce the effect of unimportant features. A Lasso model also can be used as a baseline model to compare the linear and nonlinear effects of input features. We set the $L_1$ coefficient of our Lasso model to be 0.05.

*Ridge*[60]. Ridge regression is similar to the Lasso model except it uses $L_2$ norm as a penalty to reduce the effect of unimportant features. We set the $L_2$ coefficient of our Ridge model to be 0.05.

**Training details.** YieldNet network was trained in an end-to-end manner. We initialized the network parameters with Xavier initialization[61]. To minimize the loss function defined in Eq. (1), we used Adam optimizer[62] with a learning rate of 0.05% and a mini-batch size of 32. The network was trained 4000 iterations to convergence. We do not use dropout[63] because batch normalization also has a regularization effect on the network.

**Final results.** After having trained all models, we evaluated the performances of our proposed method along with other competing models to predict corn and soybean yields. To completely evaluate all models, we took 3 years 2016, 2017, and 2018 as test years and predicted corn and soybean yields 4 times a year during the growing season on the 23rd day of July, August, September, and October. From a practical perspective, monitoring crop yield throughout the growing season is vital for optimal farm management practices (when to add fertilizer, apply pesticides, irrigate crops, etc). In addition, crop yield monitoring affects crop commodity market, which determines the future prices of crops. Therefore, our results emulate the situation where we are progressively predicting yield at the end of the growing season months earlier as a means of tracking expected crop growth. Tables 4 and 5 present the corn and soybean yield prediction results, respectively, and compare the performances of models with respect to the root–mean–square error (RMSE) evaluation metric which is defined as follows:

| Test date | | Models | | | | | | |
|---|---|---|---|---|---|---|---|---|
| Year | Month | Ridge | Lasso | RF | DFNN | RT | 3D-CNN | YieldNet |
| 2016 | July | 23.12 | 21.03 | 22.48 | 22.16 | 29.41 | 18.84 | **18.73** |
| | August | 23.16 | 19.68 | 20.95 | 20.48 | 29.16 | **15.25** | 15.76 |
| | September | 24.53 | 20.6 | 21.23 | 21.04 | 29.31 | 16.55 | **15.96** |
| | October | 24.93 | 21.05 | 21.15 | 20.74 | 27.96 | 16.65 | **15.85** |
| 2017 | July | 30.55 | 27.53 | 26.61 | 26.40 | 33.64 | 22.50 | **20.88** |
| | August | 25.16 | 22.27 | 22.25 | 20.85 | 28.02 | **16.60** | 17.74 |
| | September | 24.15 | 21.5 | 21.99 | 19.21 | 26.8 | 15.71 | **15.53** |
| | October | 25.73 | 20.94 | 22.14 | 18.90 | 26.78 | 15.69 | **15.40** |
| 2018 | July | 27.51 | 21.21 | 22.38 | 22.85 | 27.69 | **20.64** | 22.08 |
| | August | 24.5 | 19.46 | 21.52 | 21.14 | 29.34 | 18.81 | **18.25** |
| | September | 25.1 | 18.69 | 21.7 | 20.57 | 28.91 | 17.58 | **16.89** |
| | October | 32.5 | 19.2 | 22.28 | 21.63 | 28.9 | 17.72 | **16.75** |
| Average | | 25.91 | 21.10 | 22.22 | 21.33 | 28.83 | 17.71 | **17.49** |

**Table 4.** The RMSE of corn yield prediction performance of models. The average ± standard deviation for corn yield in years 2016, 2017, and 2018 are, respectively, $165.72 \pm 30.35$, $168.50 \pm 32.88$, and $170.77 \pm 34.95$. The numbers of test samples in years 2016, 2017, 2018 for corn yield prediction are 885, 882, and 784, respectively. The unit of RMSE is bushels per acre. Best results are given in bold.

| Test date | | Models | | | | | | |
|---|---|---|---|---|---|---|---|---|
| Year | Month | Ridge | Lasso | RF | DFNN | RT | 3D-CNN | YieldNet |
| 2016 | July | 9.87 | 9.12 | 9.38 | 8.27 | 10.82 | 7.08 | **5.43** |
| | August | 8.2 | 8.42 | 9.15 | 6.89 | 10.65 | 5.36 | **4.59** |
| | September | 8.3 | 8.27 | 9.12 | 7.19 | 10.37 | 4.8 | **4.45** |
| | October | 8.49 | 8.85 | 9.02 | 7.62 | 10.7 | 4.9 | **4.24** |
| 2017 | July | 8.81 | 7.01 | 6.62 | 6.30 | 9.63 | **5.66** | 5.83 |
| | August | 6.77 | 6.11 | 5.65 | 5.42 | 7.85 | **4.61** | 5.11 |
| | September | 6.82 | 6.44 | 5.62 | 5.57 | 7.88 | **4.42** | 4.55 |
| | October | 6.74 | 6.63 | 5.67 | 5.40 | 7.79 | 4.38 | **4.35** |
| 2018 | July | 8.77 | 9.88 | 8.13 | 8.62 | 9.74 | 6.58 | **6.36** |
| | August | 7.75 | 8.08 | 7.78 | 7.14 | 9.58 | 5.8 | **5.59** |
| | September | 8.25 | 8.04 | 7.78 | 7.53 | 9.64 | 5.67 | **5.16** |
| | October | 8.01 | 8.14 | 7.9 | 7.41 | 9.4 | 5.41 | **5.19** |
| Average | | 8.07 | 7.92 | 7.65 | 6.95 | 9.50 | 5.38 | **5.07** |

**Table 5.** The RMSE of soybean yield prediction performance of models. The average ± standard deviation for corn yield in years 2016, 2017, and 2018 are, respectively, $53.94 \pm 7.23$, $50.24 \pm 8.72$, and $53.17 \pm 9.72$. The numbers of test samples in years 2016, 2017, and 2018 for soybean yield prediction are 773, 776, and 663, respectively. The unit of RMSE is bushels per acre. Best results are given in bold.

$$RMSE = \sqrt{\frac{1}{N}\sum_{i=1}^{N}(y_i - \hat{y}_i)^2} \quad (2)$$

where, $N$, $y_i$, and $\hat{y}_i$ denote the number of observations, the predicted yield for $i$th observation, and the ground truth yield for $i$th observation, respectively.

As shown in Tables 4 and 5, our proposed method outperforms other methods to varying extents. The Ridge and Lasso had comparable performances for soybean yield prediction, but, Lasso performed considerably better compared to the Ridge for corn yield prediction. DFNN showed a better performance than RF, RT, and Ridge for both corn and soybean yield predictions. DFNN had similar performance with Lasso for corn yield prediction while having higher prediction accuracy for soybean yield prediction. Despite the linear modeling structure, Lasso performed better than RF and RT for corn yield prediction, which indicates that RF and RT cannot successfully capture the nonlinearity of remote sensing data, resulting in poor performance compared to Lasso. RT had a weak performance compared to other methods due to being prone to overfitting. RF performs considerably better than RT because of using ensemble learning, which makes it robust against overfitting. 3D-CNN outperformed other models except for our proposed method due to capturing the temporal effects and spatial and intra-band features of remote sensing data. 3D-CNN performed slightly better than YieldNet in three cases for soybean and corn yield predictions. But, YieldNet outperformed the 3D-CNN on average for both corn and soybean yield predictions while having a smaller number of parameters and computation time.

| Measures | Models | | | | | | |
|---|---|---|---|---|---|---|---|
| | Ridge | Lasso | RF | DFNN | RT | 3D-CNN | YieldNet |
| Number of parameters | 8641 | 8641 | Non-parametric | 923,882 | Non-parametric | 3,167,490 | 1,436,050 |
| Training time (min) | 0.42 | 1.94 | 155.14 | 31.46 | 1.44 | 759.4 | 2.89 |

**Table 6.** The number of parameters and training time of the competing methods for crop yield prediction. The training time indicates the total time needed for training models for both corn and soybean yield prediction. The models were trained on an Intel i7-4790 CPU 3.60 GHz.

| Crop | 2016 | | | | 2017 | | | | 2018 | | | |
|---|---|---|---|---|---|---|---|---|---|---|---|---|
| | Jul. | Aug. | Sep. | Oct. | Jul. | Aug. | Sep. | Oct. | Jul. | Aug. | Sep. | Oct. |
| Corn | 14.92 | 14.32 | 14.36 | 14.48 | 18.24 | 15.92 | 12.71 | 13.40 | 16.11 | 14.59 | 13.35 | 13.23 |
| Soybean | 6.05 | 4.98 | 4.10 | 3.66 | 5.0 | 5.13 | 3.73 | 4.11 | 5.18 | 4.90 | 3.91 | 4.19 |

**Table 7.** The MAE of the corn and soybean yield prediction performances of our proposed model. The unit of MAE is bushels per acre.

Our proposed method outperformed the other methods due to multiple factors: (1) the convolution operation in the YieldNet model captures both the temporal effect of remote sensing data collected over growing season and the spatial information of bins in histograms, (2) the YieldNet network uses transfer learning between corn and soybean yield predictions by sharing the weights of the backbone feature extractor, and (3) using a shared backbone feature extractor in the YieldNet model substantially decreases the number of model's parameters and subsequently helps training process despite having the limited labeled data. The prediction accuracy decreases as we try to make predictions earlier during the growing season (e.g. July and August) due to loss of information. All models except our proposed model do not show a clear decreasing pattern in performance accuracy as we go from October to July, which indicates they cannot fully learn the functional mapping from satellite images to the yield.

Table 6 compares the number of parameters and training time of the competing methods for crop yield prediction. As shown in Table 6, Lasso and Ridge have the lowest number of parameters compared to neural network based models such as DFNN, 3D-CNN, and YieldNet due to their linear modeling structure. Compared to the YieldNet model, other models should be trained separately for corn and soybean, which results in having twice the total number of parameters. 3D-CNN has the highest number of parameters among the neural network based models. From a computation time perspective, linear models and RT had the lowest training time. RF had the second longest training time after 3D-CNN compared to other models due to using ensemble learning. Among neural network based models, YieldNet had the shortest training time and 3D-CNN had the longest training time. The models were trained on an Intel i7-4790 CPU 3.60 GHz. The inference times of all methods are less than a second. However, the inference time of our model is less than that of the 3D-CNN model.

We also report the yield prediction performance of our proposed model with respect to another evaluation metric, mean absolute error (MAE), in Table 7. As shown in Table 7, our proposed method accurately predicted corn yield 1 month, 2 months, 3 months, and 4 months before the harvest with MAE being 9.92%, 8.88%, 8.36%, and 7.8% of the average corn yield, respectively. The proposed model also accurately predicted soybean yield one month, two months, three months, and four months before harvest with MAE being 10.05%, 9.06%, 8.01%, and 7.67% of the average soybean yield, respectively. The proposed model is slightly more accurate in soybean yield forecasting than corn yield forecasting, which is due to the higher variation in the corn yield compared to the soybean yield.

We visualized the error percentage maps for the corn and soybean yield predictions for the year 2018. As shown in Figs. 5 and 6, the error percentage is below 5% for most counties, which indicates that our proposed model provides a robust and accurate yield prediction across US Corn Belt.

To further evaluate the prediction results of our proposed model, we created the scatter plots of ground truth yield against the predicted yield for the year 2018. Figure 7 depicts the scatter plots for the corn yield prediction during the growing season in the months July, August, September, and October. The corn scatter plots indicate that the YieldNet model can successfully forecast yield months prior to harvesting.

Figure 8 depicts the scatter plots for the soybean yield prediction during the growing season in the months July, August, September, and October. The corn scatter plots indicate that the YieldNet model provides reliable and accurate yield months prior to the harvest.

### Ablation study
In order to examine the usefulness of using a single deep learning model for simultaneously predicting the yield of two crops, we perform the following analysis. We train two separate models one for corn yield prediction and another for soybean yield prediction which are as follows:

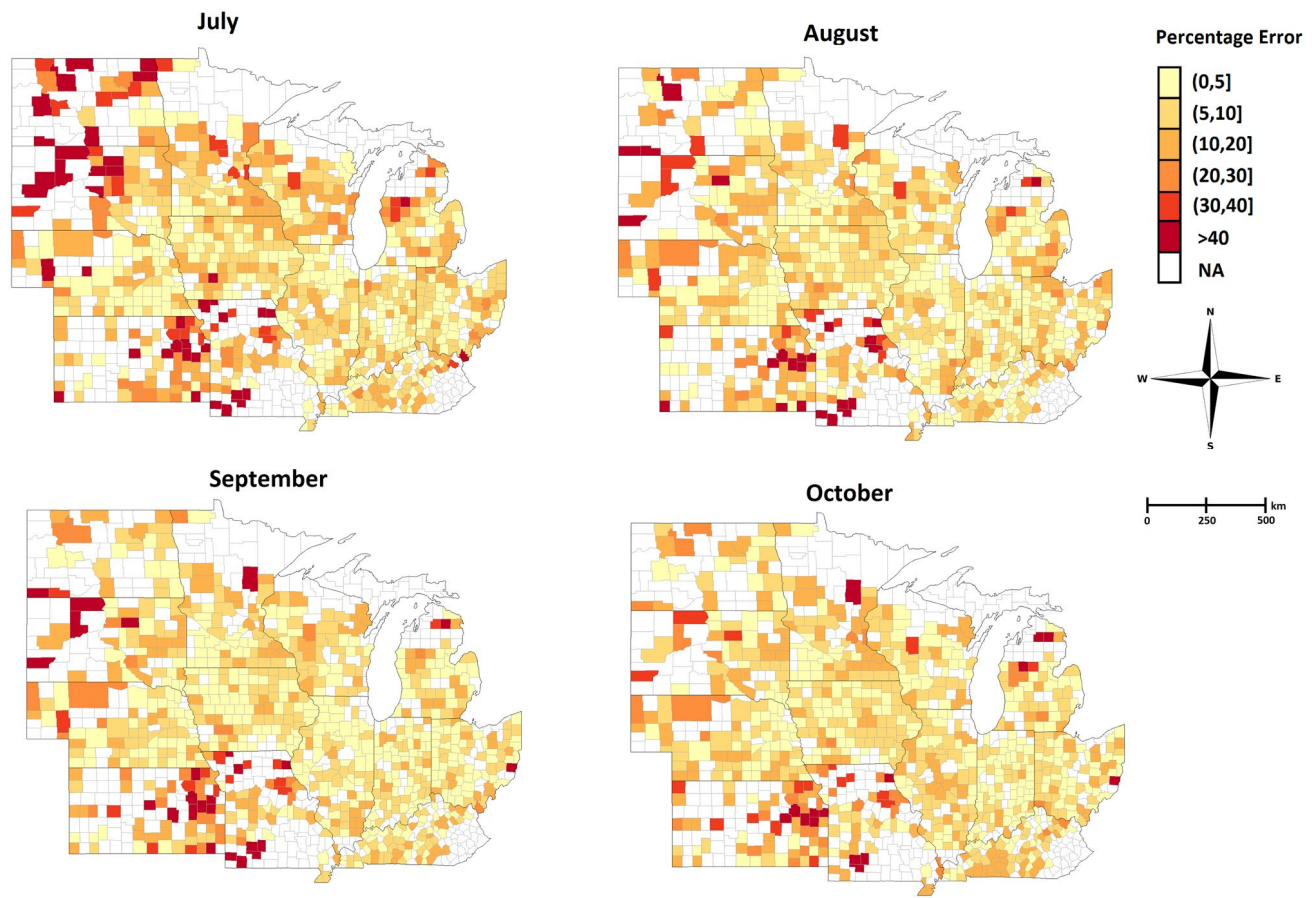

**Figure 5.** The error percentage maps for the 2018 corn yield prediction which is done during growing season in July, August, September, and October. The counties with white color indicate the ground truth yield were not available for those counties in 2018.

*YieldNet$^{corn}$* This model has exactly the same network architecture as the YieldNet model except we removed the soybean head from the original YieldNet network. As a result, *YieldNet$^{corn}$* can only predict the corn yield.

*YieldNet$^{soy}$* This model has exactly the same network architecture as the YieldNet model except we removed the corn head from the original YieldNet network. As a result, *YieldNet$^{soy}$* can only predict the soybean yield.

Tables 8 and 9 compare the yield prediction performances of the above-mentioned models with the original YieldNet model.

As shown in Tables 8 and 9, the YieldNet model which simultaneously predicts corn and soybean yields outperforms individual *YieldNet$^{corn}$* and *YieldNet$^{soy}$* models. The YieldNet provides more robust and accurate yield predictions compared to the other two individual models, which indicates that transfer learning between corn and soybean yield prediction improves the yield prediction accuracy for both crops.

## Discussion and conclusion

Our numerical results illustrate that our approach to simultaneously predicting yield for both corn and soybeans is possible and can achieve higher accuracy than individual models. By utilizing transfer learning between corn and soybean yield to share the weights of the backbone feature extractor, YieldNet was able to substantially decrease the number of learning parameters. Our transfer learning approach enabled us to save on computation resources while also maximizing our prediction accuracy. Moreover, the accuracy achieved using a 4-month look-ahead has a lot of important implications for crop management decisions. With accurate yield predictions at various time points, decision-makers now have the ability to change crop management practices to ensure yield is being maximized throughout its growth stage.

Although our approach highlighted corn and soybean in the US market, this approach is applicable to any number of crops in any region. Due to the strength of our deep learning framework in combination with a generalized loss function, our approach is ready for scale. To improve the accuracy of our methodology, more data can be gathered and this can be left as a future extension to this work, alongside more crops and more regions. It is the hope of this paper that our approach and results showcase the power deep learning for simultaneous yield prediction can have on the remote sensing community and the larger agricultural community as a whole.

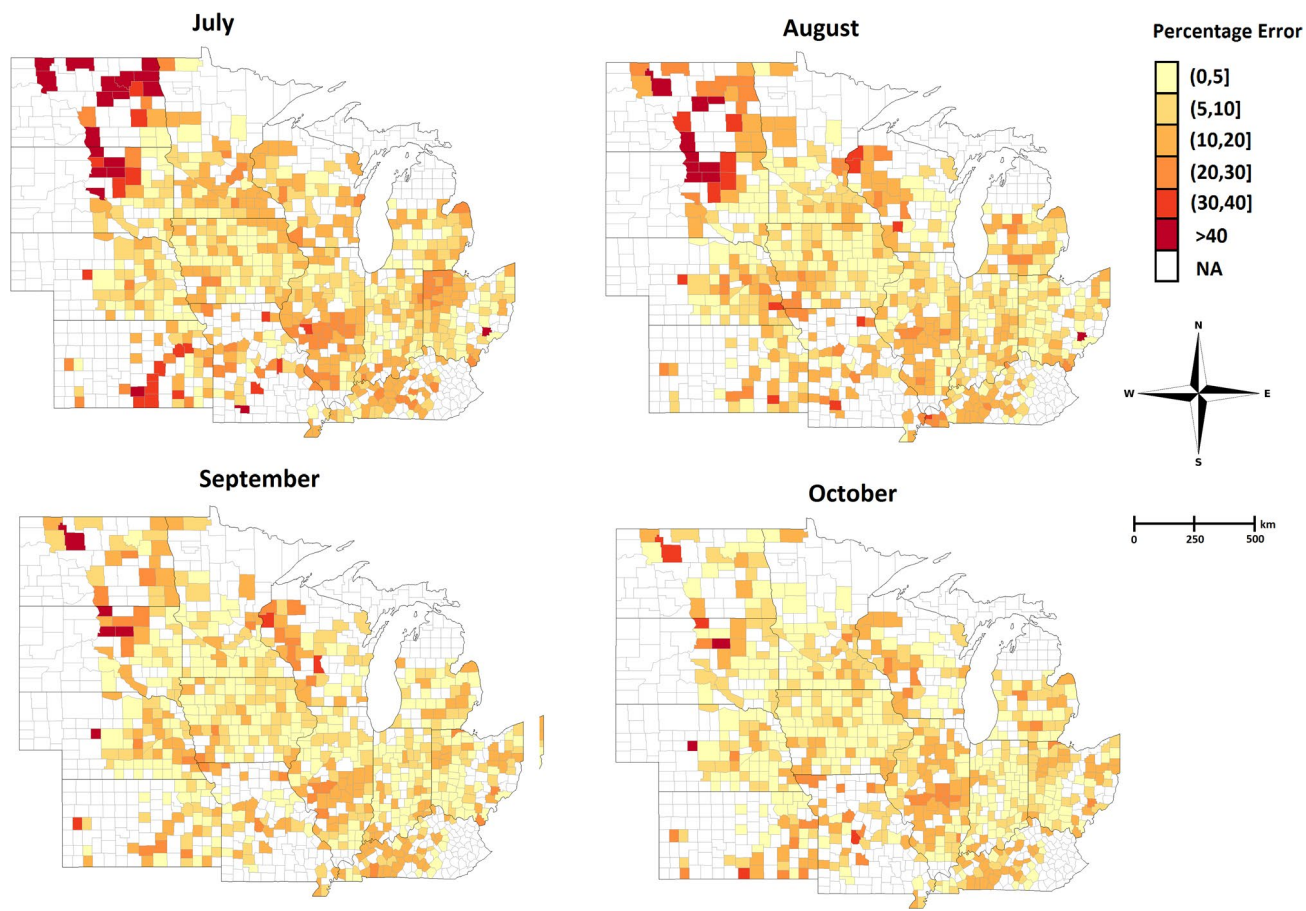

**Figure 6.** The error percentage maps for the 2018 soybean yield prediction which is done during the growing season in July, August, September, and October. The counties with white color indicate the ground truth yield were not available for those counties in 2018.

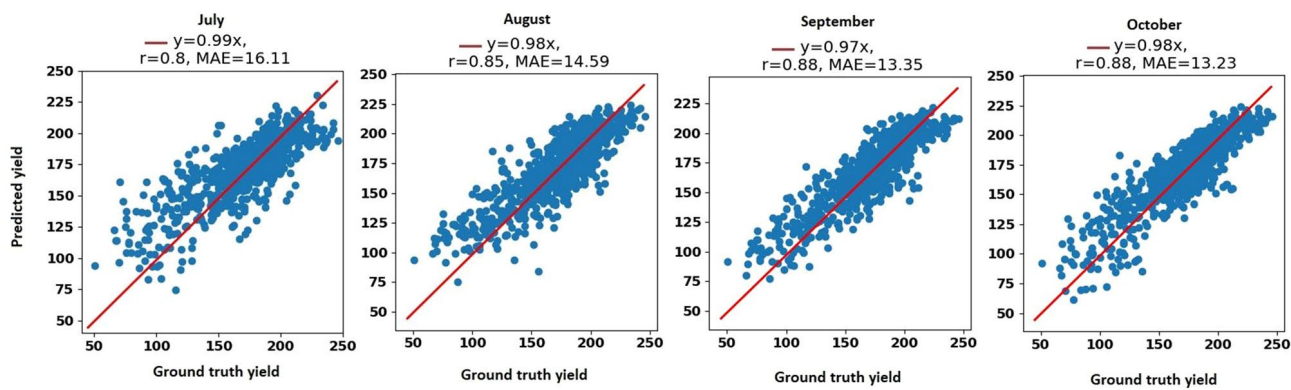

**Figure 7.** The scatter plots for the 2018 corn yield prediction during the growing season in the months July, August, September, and October. MAE and r stand for mean absolute error and correlation coefficient, respectively. The unit of yield is bushels per acre.

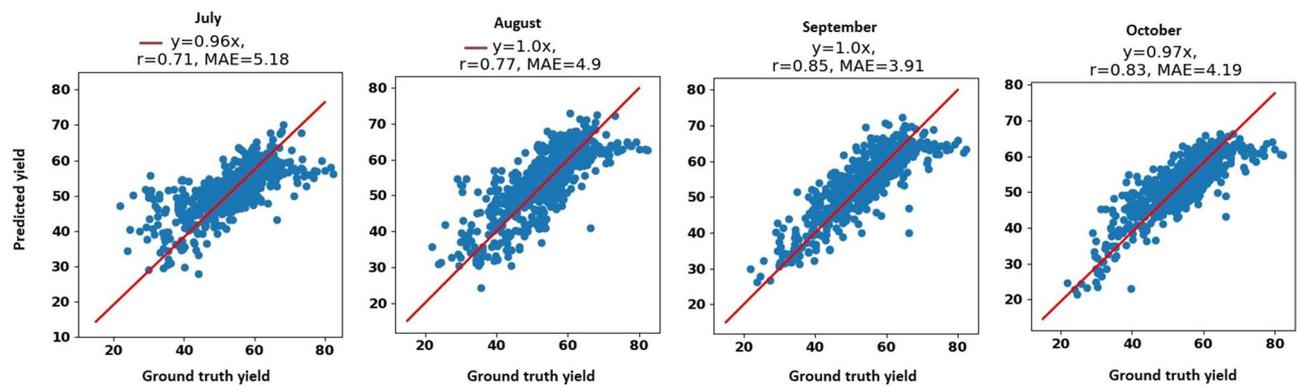

**Figure 8.** The scatter plots for the 2018 soybean yield prediction during the growing season in months July, August, September, and October. MAE and r stand for mean absolute error and correlation coefficient, respectively. The unit of yield is bushels per acre.

| | 2016 | | | | 2017 | | | | 2018 | | | |
|---|---|---|---|---|---|---|---|---|---|---|---|---|
| Model | Jul. | Aug. | Sep. | Oct. | Jul. | Aug. | Sep. | Oct. | Jul. | Aug. | Sep. | Oct. |
| $YieldNet^{corn}$ | 21.74 | 18.88 | 17.55 | 16.90 | 23.80 | 18.84 | 16.83 | 16.83 | 23.71 | 18.60 | 17.84 | 17.62 |
| YieldNet | 18.73 | 15.76 | 15.96 | 15.85 | 20.88 | 17.74 | 15.53 | 15.40 | 22.08 | 18.25 | 16.89 | 16.75 |

**Table 8.** The RMSE of the corn yield prediction performances of the YieldNet and $YieldNet^{corn}$ models. The unit of RMSE is bushels per acre.

| | 2016 | | | | 2017 | | | | 2018 | | | |
|---|---|---|---|---|---|---|---|---|---|---|---|---|
| Model | Jul. | Aug. | Sep. | Oct. | Jul. | Aug. | Sep. | Oct. | Jul. | Aug. | Sep. | Oct. |
| $YieldNet^{soy}$ | 6.49 | 5.49 | 5.15 | 4.57 | 6.96 | 5.82 | 4.86 | 4.86 | 7.80 | 6.58 | 5.46 | 5.57 |
| YieldNet | 5.43 | 4.59 | 4.45 | 4.24 | 5.83 | 5.11 | 4.55 | 4.35 | 6.36 | 5.59 | 5.16 | 5.19 |

**Table 9.** The RMSE of the soybean yield prediction performances of the YieldNet and $YieldNet^{soy}$ models. The unit of yield is bushels per acre.

### Data availability
This data is publicly available and acquisition details are in "Data".

### References
 1. Anastasiou, E. et al. Satellite and proximal sensing to estimate the yield and quality of table grapes. *Agriculture* **8**, 94 (2018).
 2. da Silva, C. A., Nanni, M. R., Teodoro, P. E. & Silva, G. F. C. Vegetation indices for discrimination of soybean areas: A new approach. *Agron. J.* **109**, 1331–1343 (2017).
 3. Quarmby, N., Milnes, M., Hindle, T. & Silleos, N. The use of multi-temporal ndvi measurements from avhrr data for crop yield estimation and prediction. *Int. J. Remote. Sens.* **14**, 199–210 (1993).
 4. Kogan, F., Gitelson, A., Zakarin, E., Spivak, L. & Lebed, L. Avhrr-based spectral vegetation index for quantitative assessment of vegetation state and productivity. *Photogramm. Eng. Remote. Sens.* **69**, 899–906 (2003).
 5. Singh, R. P., Roy, S. & Kogan, F. Vegetation and temperature condition indices from noaa avhrr data for drought monitoring over india. *Int. J. Remote Sensing* **24**, 4393–4402 (2003).
 6. Liou, Y.-A. & Kar, S. K. Evapotranspiration estimation with remote sensing and various surface energy balance algorithms—A review. *Energies* **7**, 2821–2849 (2014).
 7. Song, L., Liu, S., Kustas, W. P., Zhou, J. & Ma, Y. Using the surface temperature-albedo space to separate regional soil and vegetation temperatures from aster data. *Remote. Sens.* **7**, 5828–5848 (2015).
 8. Geipel, J., Link, J. & Claupein, W. Combined spectral and spatial modeling of corn yield based on aerial images and crop surface models acquired with an unmanned aircraft system. *Remote. Sens.* **6**, 10335–10355 (2014).
 9. Van Wart, J., Kersebaum, K. C., Peng, S., Milner, M. & Cassman, K. G. Estimating crop yield potential at regional to national scales. *Field Crop. Res.* **143**, 34–43 (2013).
 10. Battude, M. et al. Estimating maize biomass and yield over large areas using high spatial and temporal resolution sentinel-2 like remote sensing data. *Remote. Sens. Environ.* **184**, 668–681 (2016).
 11. Fieuzal, R., Sicre, C. M. & Baup, F. Estimation of corn yield using multi-temporal optical and radar satellite data and artificial neural networks. *Int. J. Appl. Earth Observ. Geoinf.* **57**, 14–23 (2017).

12. Sagan, V. *et al.* Uav-based high resolution thermal imaging for vegetation monitoring, and plant phenotyping using ici 8640 p, flir vue pro r 640, and thermomap cameras. *Remote. Sens.* **11**, 330 (2019).
13. Sellam, V. & Poovammal, E. Prediction of crop yield using regression analysis. *Indian J. Sci. Technol.* **9**, 1–5 (2016).
14. Shahhosseini, M., Martinez-Feria, R. A., Hu, G. & Archontoulis, S. V. Maize yield and nitrate loss prediction with machine learning algorithms. *Environ. Res. Lett.* **14**, 124026 (2019).
15. Suresh, G., Kumar, A. S., Lekashri, S. & Manikandan, R. Efficient crop yield recommendation system using machine learning for digital farming. *Int. J. Mod. Agric.* **10**, 906–914 (2021).
16. Chu, Z. & Yu, J. An end-to-end model for rice yield prediction using deep learning fusion. *Comput. Electron. Agric.* **174**, 105471 (2020).
17. Nassar, L., Okwuchi, I. E., Saad, M., Karray, F., Ponnambalam, K., & Agrawal, P. Prediction of strawberry yield and farm price utilizing deep learning. In *2020 International Joint Conference on Neural Networks (IJCNN)*, 1–7 (IEEE, 2020).
18. Bhojani, S. H. & Bhatt, N. Wheat crop yield prediction using new activation functions in neural network. *Neural. Comput. Appl.* 1–11 (2020).
19. Khaki, S., Wang, L. & Archontoulis, S. V. A cnn-rnn framework for crop yield prediction. *Front. Plant Sci.* **10**, 1750 (2020).
20. Chang, A., Jung, J., Yeom, J., Maeda, M. M., Landivar, J. A., Enciso, J. M., Avila, C. A. & Anciso, J. R. Unmanned Aircraft System-(UAS-) Based High-Throughput Phenotyping (HTP) for Tomato Yield Estimation. *J. Sens.* (2021).
21. Zhou, J. *et al.* Yield estimation of soybean breeding lines under drought stress using unmanned aerial vehicle-based imagery and convolutional neural network. *Biosyst. Eng.* **204**, 90–103 (2021).
22. Apolo-Apolo, O., Martínez-Guanter, J., Egea, G., Raja, P. & Pérez-Ruiz, M. Deep learning techniques for estimation of the yield and size of citrus fruits using a uav. *Eur. J. Agron.* **115**, 126030 (2020).
23. Rischbeck, P. *et al.* Data fusion of spectral, thermal and canopy height parameters for improved yield prediction of drought stressed spring barley. *Eur. J. Agron.* **78**, 44–59 (2016).
24. Kuwata, K. & Shibasaki, R. Estimating corn yield in the united states with modis evi and machine learning methods. *ISPRS Ann. Photogramm. Remote Sens. Spat. Inf. Sci.* **3**(8), 131–136 (2016).
25. Leroux, L. *et al.* Maize yield estimation in west africa from crop process-induced combinations of multi-domain remote sensing indices. *Eur. J. Agron.* **108**, 11–26 (2019).
26. Gómez, D., Salvador, P., Sanz, J. & Casanova, J. L. Potato yield prediction using machine learning techniques and sentinel 2 data. *Remote. Sens.* **11**, 1745 (2019).
27. Zhuo, W. *et al.* Assimilating soil moisture retrieved from sentinel-1 and sentinel-2 data into wofost model to improve winter wheat yield estimation. *Remote. Sens.* **11**, 1618 (2019).
28. Awad, M. M. Toward precision in crop yield estimation using remote sensing and optimization techniques. *Agriculture* **9**, 54 (2019).
29. Ballesteros, R., Intrigliolo, D. S., Ortega, J. F., Ramírez-Cuesta, J. M., Buesa, I. & Moreno, M. A. (2020). Vineyard yield estimation by combining remote sensing, computer vision and artificial neural network techniques. *Precis. Agric.* **21**, 1242–1262 (2020).
30. Wang, Y., Zhang, Z., Feng, L., Du, Q. & Runge, T. Combining multi-source data and machine learning approaches to predict winter wheat yield in the conterminous united states. *Remote. Sens.* **12**, 1232 (2020).
31. Maimaitijiang, M. *et al.* Soybean yield prediction from uav using multimodal data fusion and deep learning. *Remote. Sens. Environ.* **237**, 111599 (2020).
32. Cao, J. *et al.* Integrating multi-source data for rice yield prediction across china using machine learning and deep learning approaches. *Agric. For. Meteorol.* **297**, 108275 (2021).
33. Paudel, D. *et al.* Machine learning for large-scale crop yield forecasting. *Agric. Syst.* **187**, 103016 (2021).
34. Sun, J. *et al.* Multilevel deep learning network for county-level corn yield estimation in the us corn belt. *IEEE J. Sel. Top. Appl. Earth Obs. Remote. Sens.* **13**, 5048–5060 (2020).
35. USDA. USDA long-term agricultural projections. https://www.usda.gov/oce/commodity/projection (2019).
36. Jin, X. *et al.* A 315 review of data assimilation of remote sensing and crop models. *Eur. J. Agron.* **92**, 141–152 (2018).
37. Zhu, B. *et al.* A regional maize yield hierarchical linear model combining landsat 8 vegetative indices and meteorological data: Case study in jilin province. *Remote. Sens.* **13**, 356 (2021).
38. Tuia, D., Verrelst, J., Alonso, L., Pérez-Cruz, F. & Camps-Valls, G. Multioutput support vector regression for remote sensing bio-physical parameter estimation. *IEEE Geosci. Remote. Sens. Lett.* **8**, 804–808 (2011).
39. Alebele, Y. *et al.* Estimation of canopy biomass components in paddy rice from combined optical and sar data using multi-target gaussian regressor stacking. *Remote. Sens.* **12**, 2564 (2020).
40. Santana, E. J. *et al.* Predicting poultry meat characteristics using an enhanced multi-target regression method. *Biosyst. Eng.* **171**, 193–204 (2018).
41. da Silva, B. L. S., Inaba, F. K., Salles, E. O. T. & Ciarelli, P. M. Outlier robust extreme machine learning for multi-target regression. *Expert. Syst. Appl.* **140**, 112877 (2020).
42. Xiao, X. & Xu, Y. Multi-target regression via self-parameterized Lasso and refactored target space. *Appl. Intell.* 1–9 (2021).
43. Deng, J., Dong, W., Socher, R., Li, L. J., Li, K. & Fei-Fei, L. Imagenet: A large-scale hierarchical image database. In *2009 IEEE conference on computer vision and pattern recognition*, 248–255 (IEEE, 2009).
44. You, J., Li, X., Low, M., Lobell, D. & Ermon, S. Deep gaussian process for crop yield prediction based on remote sensing data. In *Thirty-First AAAI Conference on Artificial Intelligence* (2017).
45. LeCun, Y., Bengio, Y. & Hinton, G. Deep learning. *Nature* **521**, 436–444 (2015).
46. Goodfellow, I., Bengio, Y., Courville, A. & Bengio, Y. *Deep Learning* Vol. 1 (MIT Press Cambridge, 2016).
47. Ioffe, S. & Szegedy, C. Batch normalization: Accelerating deep network training by reducing internal covariate shift. arXiv preprint arXiv:1502.03167 (2015).
48. Abadi, M. *et al.* Tensorflow: A system for large-scale machine learning. In *12th {USENIX} Symposium on Operating Systems Design and Implementation ({OSDI} 16)*, 265–283 (2016).
49. Usda—National agricultural statistics service quickstats. https://quickstats.nass.usda.gov/. (Accessed 12–30, 2020).
50. Vermote, E. MOD09A1 MODIS/terra surface reflectance 8-day l3 global 500m sin grid v006. NASA EOSDIS Land Process. DAAC https://doi.org/10.5067/MODIS/MOD09A1.006 (2015).
51. Wan, Z., Hook, S. & Hulley, G. MOD11A2 MODIS/Terra land surface temperature/emissivity 8-day l3 global 1km sin grid v006. NASA EOSDIS Land Process. DAAC https://doi.org/10.5067/MODIS/MOD11A2.006 (2015).
52. Vermote, E. MOD09Q1 MODIS/terra surface reflectance 8-day l3 global 250m sin grid v006. NASA EOSDIS Land Process. *DAAC*.
53. USDA-NASS. USDA national agricultural statistics service cropland data layer. (2020).
54. Breiman, L. Random forests. *Mach. Learning* **45**, 5–32 (2001).
55. Tran, D., Bourdev, L., Fergus, R., Torresani, L. & Paluri, M. Learning spatiotemporal features with 3d convolutional networks. In *Proceedings of the IEEE International Conference on Computer Vision*, 4489–4497 (2015).
56. Li, Y., Zhang, H. & Shen, Q. Spectral–spatial classification of hyperspectral imagery with 3d convolutional neural network. *Remote. Sens.* **9**, 67 (2017).
57. Nevavuori, P., Narra, N., Linna, P. & Lipping, T. Crop yield prediction using multitemporal uav data and spatio-temporal deep learning models. *Remote. Sens.* **12**, 4000 (2020).
58. Breiman, L., Friedman, J., Stone, C. J. & Olshen, R. A. *Classification and Regression Trees* (CRC Press, 1984).
59. Tibshirani, R. Regression shrinkage and selection via the lasso. *J. R. Stat. Soc. Ser. B (Methodol.)* **58**, 267–288 (1996).

60. Hoerl, A. E. & Kennard, R. W. Ridge regression: Biased estimation for nonorthogonal problems. *Technometrics* **12**, 55–67 (1970).
61. Glorot, X. & Bengio, Y. Understanding the difficulty of training deep feedforward neural networks. In *Proceedings of the Thirteenth International Conference on Artificial Intelligence and Statistics*, 249–256 (2010).
62. Kingma, D. P. & Ba, J. Adam: A method for stochastic optimization. arXiv preprint arXiv:1412.6980 (2014).
63. Srivastava, N., Hinton, G., Krizhevsky, A., Sutskever, I. & Salakhutdinov, R. Dropout: A simple way to prevent neural networks from overfitting. *J. Mach. Learn. Res.* **15**, 1929–1958 (2014).

### Author contributions
S.K., H.P., and L.W. conceived the study. S.K. implemented the computational experiments. S.K., H.P., and L.W. wrote the paper.

### Funding
This work was partially supported by the National Science Foundation under the LEAP HI and GOALI programs (Grant number 1830478) and under the EAGER program (Grant number 1842097).

### Competing interests
The authors declare no competing interests.

### Additional information
**Correspondence** and requests for materials should be addressed to S.K.

**Reprints and permissions information** is available at www.nature.com/reprints.

**Publisher's note**  Springer Nature remains neutral with regard to jurisdictional claims in published maps and institutional affiliations.